\documentclass[conference]{IEEEtran}
\IEEEoverridecommandlockouts
\usepackage{cite}
\usepackage{amsmath,amssymb,amsfonts}
\usepackage{algorithmic}
\usepackage{graphicx}
\usepackage{subcaption}
\usepackage{textcomp}
\usepackage{xcolor}
\def\BibTeX{{\rm B\kern-.05em{\sc i\kern-.025em b}\kern-.08em
    T\kern-.1667em\lower.7ex\hbox{E}\kern-.125emX}}
\begin{document}

\title{Precipitation Nowcasting Using Physics Informed Discriminator Generative Models}

\author{
\IEEEauthorblockN{
Junzhe Yin\textsuperscript{1},
Cristian Meo\textsuperscript{1},
Ankush Roy\textsuperscript{1}, 
Zeineb Bou Cher\textsuperscript{1}, 
Yanbo Wang\textsuperscript{1},
Ruben Imhoff\textsuperscript{2}}
\IEEEauthorblockN{
Remko Uijlenhoet\textsuperscript{1}, 
Justin Dauwels\textsuperscript{1}
}

\IEEEauthorblockA{\textsuperscript{1}Delft University of Technology, Netherlands}
\IEEEauthorblockA{\textsuperscript{2}Deltares, Netherlands}
}

\author{
\IEEEauthorblockN{
Junzhe Yin\textsuperscript{1*},
Cristian Meo\textsuperscript{1*},
Ankush Roy\textsuperscript{1}, 
Zeineh Bou Cher\textsuperscript{1}, 
Mircea Lică\textsuperscript{1}, 
Yanbo Wang\textsuperscript{1},
Ruben Imhoff\textsuperscript{2}}
\IEEEauthorblockN{
Remko Uijlenhoet\textsuperscript{1}, 
Justin Dauwels\textsuperscript{1}
}

\IEEEauthorblockA{\textsuperscript{1}Delft University of Technology, Netherlands\\
\textsuperscript{2}Deltares, Netherlands}

\IEEEauthorblockA{\textsuperscript{*}Equal Contribution}
\IEEEauthorblockA{Corrisponding author: Junzhe.Yin1999@gmail.com}
}

\maketitle

\begin{abstract}
Nowcasting leverages real-time atmospheric conditions to forecast weather over short periods. State-of-the-art models, including PySTEPS, encounter difficulties in accurately forecasting extreme weather events because of their unpredictable distribution patterns. In this study, we design a physics-informed neural network to perform precipitation nowcasting using the precipitation and meteorological data from the Royal Netherlands Meteorological Institute (KNMI). This model draws inspiration from the novel Physics-Informed Discriminator GAN (PID-GAN) formulation, directly integrating physics-based supervision within the adversarial learning framework. The proposed model adopts a GAN structure, featuring a Vector Quantization Generative Adversarial Network (VQ-GAN) and a Transformer as the generator, with a temporal discriminator serving as the discriminator. Our findings demonstrate that the PID-GAN model outperforms numerical and SOTA deep generative models in terms of precipitation nowcasting downstream metrics.
\end{abstract}

\section{INTRODUCTION}
\footnote{Accepted at European conference on signal processing (EUSIPCO) 2024}
More frequent global extreme precipitation has led to severe flooding, soil erosion, loss of agricultural productivity, and heightened health risks\cite{tabari2020climate}. Traditional Numerical Weather Prediction (NWP) models, though comprehensive, face challenges in short-term forecasting of rainfall due to their high computational requirements and too low spatial-temporal resolution \cite{prudden2020review, imhoff2020spatial}. Precipitation nowcasting techniques aim to provide accurate forecasts of upcoming precipitation events within six hours for local regions, reducing response time and efforts to handle extreme weather events \cite{ravuri2021skilful, Zhang2023Skillful, Bi2023Accurate}. Radar extrapolation methods, such as PySteps, leverage real-time data and use optical flow and statistical analysis techniques to improve weather prediction accuracy \cite{pulkkinen2019pysteps}. However, while statistical nowcasting methods eliminate the need for historical data, they do not explicitly account for the growth and decay processes of convective rainfall, limiting their usefulness for severe rainfall events. Due to limitations in radar extrapolation methods, there has been a shift toward deep learning models \cite{ravuri2021skilful, bi2023nowcasting}, which have shown promise in weather prediction by capturing complex spatio-temporal patterns without relying on traditional data assimilation techniques. 

Deep generative models like Generative Adversarial Networks (GAN) \cite{goodfellow2014generative}  and Variational Autoencoder (VAE) \cite{kingma2013auto} have excelled in precipitation nowcasting, offering more accurate and realistic forecasts by modelling the distribution underlying the precipitation dynamics \cite{jing2019aenn,ravuri2021skilful,bi2023nowcasting}. Nevertheless, these models have many limitations, for instance, they are not able to produce consistent prediction over medium- to long-term horizons and struggle with extreme event modeling and prediction. Moreover, a notable limitation of these models is their tendency to overlook fundamental physical laws, resulting in predictions that may not always align with them, particularly in scenarios not encountered during training. Physics-informed machine learning (PIML) \cite{willard2020integrating} has emerged as a promising solution, enhancing weather and climate forecasting by improving physical consistency \cite{kashinath2021physics}. In this paper, we propose a novel PIML model that integrates deep generative models with physical priors of precipitation, aiming to produce accurate and physically consistent precipitation forecasts.

\subsection{Dataset}
This paper explores precipitation nowcasting in the Netherlands, using weather radar data from the Royal Netherlands Meteorological Institute (KNMI) for 2008-2021 \cite{bi2023nowcasting}, along with hourly meteorological data from Automatic Weather Stations (AWS) \cite{sluiter2012interpolation} and ERA5 reanalyses \cite{hersbach2020era5}. The meteorological data from AWS  contains hourly observations of wind speed, wind direction, air temperature, dew point temperature and global radiation. ERA5 offers hourly estimates for a wide variety of quantities related to the atmosphere, ocean waves, and the land surface \cite{hersbach2020era5}. We use radar precipitation maps with a 5-minute temporal resolution and a 1 km spatial resolution. Our study focuses on a $256\times256$ pixel area covering most of the country and 12 Dutch catchment regions, as detailed in [\citenum{imhoff2020spatial}, Fig.~1]. Following \cite{bi2023nowcasting}, an event is classified as extreme if its average rainfall over three hours ranks in the top 1\% of all recorded events between 2008 and 2021.
This threshold is defined as exceeding 5 mm per 3 hours, as shown in [\citenum{bi2023nowcasting}, Table.~1]. To match the temporal resolution of AWS and ERA5 meteorological data with the used radar dataset, we apply cubic interpolation to estimate half-hour intervals of the latter (e.g., \(T-60\), \(T-30\), to \(T+180\) minutes) from the original hourly data points (e.g., \(T-60\), \(T\), \(T+60\), \(T+120\), \(T+180\) minutes).
To address the limited spatial coverage of the AWS and ERA5 reanalyses datasets, we applied kriging interpolation using the PyKrige package \cite{murphy2014pykrige}, achieving a spatial resolution compatible with our radar data.
 Kriging uses spatial autocorrelation and variance to accurately estimate values in unmeasured locations, creating maps that align with the resolutions of the radar map \cite{oliver1990kriging}. 

\subsection{Problem Formulation}


In the last few years, precipitation nowcasting using deep learning models have been cast as a video prediction problem \cite{bi2023nowcasting}, where given an input spatio-temporal sequence of $N$ frames $\boldsymbol{x}_{\text{in}} \in \mathbb{R}^{N \times H \times W \times C}$, where $H, W$ denote the spatial resolution and $C$ represents the image channels or the different types of measurements (e.g., radar maps, humidity maps, etc). The goal is to predict the next $M$ precipitation maps $\boldsymbol{x}_{\text{out}} \in \mathbb{R}^{M \times H \times W \times 1}$ and classify extreme events based on the defined threshold. Throughout this work we use $N=3$ and $M=6$. 
\section{Related Works}

Generative Adversarial Networks, as the current state-of-the-art in various fields, have seen many adaptable variations in recent years, demonstrating their flexibility. They serve as key architectures in precipitation nowcasting tasks. For instance, the Adversarial Extrapolation Neural Net (AENN) \cite{jing2019aenn}, a variation of GAN with 2 discriminators, has shown superior performance in weather radar echo extrapolation than NWP approaches. In \cite{ravuri2021skilful}, a GAN utilizing temporal and spatial discriminators was proposed. This advancement underscores the potential of GANs in enhancing the accuracy of precipitation nowcasting through sophisticated data generation techniques. 

Despite their advancements, these models face persistent challenges, such as delivering consistently clear and precise forecasts and adequately generalizing across diverse weather conditions, particularly those underrepresented in training datasets. A significant obstacle faced by sophisticated neural networks such as AENN is their lack of interpretability. A critical hurdle with advanced neural networks lies in their interpretability. The uninterpretable reasoning behind their predictions complicates the understanding of their decision-making processes, making it difficult for experts to fully trust and utilize these models. Moreover, these systems must adhere to the fundamental physical principles that govern meteorological phenomena, ensuring that their predictive capabilities do not compromise physical accuracy. Maintaining a balance between physical accuracy and predictive performance represents a significant challenge, underscoring the demand for innovation in meteorological forecasting. This balance is crucial for developing reliable models that can accurately predict weather patterns while adhering to the fundamental principles of meteorology. Recently, PID-GAN, a physics-informed discriminator GAN formulation that does not suffer from an imbalance of gradient flow compared to physics-informed neural networks (PINN) \cite{raissi2019physics}, was introduced in \cite{daw2021pid}. In this work, we adopt the PID-GAN framework to integrate physical constraints into our model, ensuring the generation of physically coherent precipitation forecasts. This includes the incorporation of moisture conservation equations, as outlined in \cite{pu2019numerical}, to accurately capture the dynamics of precipitation.

\section{Methodology}
In this section, we define and describe the proposed PID-GAN architecture used in this work. Moreover, we describe the physics-based loss function used to inject the moisture conservation equation information into the PID-GAN architecture.

\subsection{PID-GAN: Model Architecture}

This section describes the backbone architecture used for the proposed PID-GAN model, which is inspired by the successful implementation detailed in \cite{ravuri2021skilful}. Our design adopts this proven structure, incorporating a generator alongside spatial and temporal discriminators, due to its demonstrated effectiveness in capturing complex spatio-temporal relationships in data. This approach ensures a robust framework capable of handling the intricate dynamics of precipitation nowcasting by efficiently learning from both the spatial patterns and temporal sequences inherent in meteorological data.

\begin{figure*}[htbp]
\centering
\includegraphics[width=0.8\textwidth]{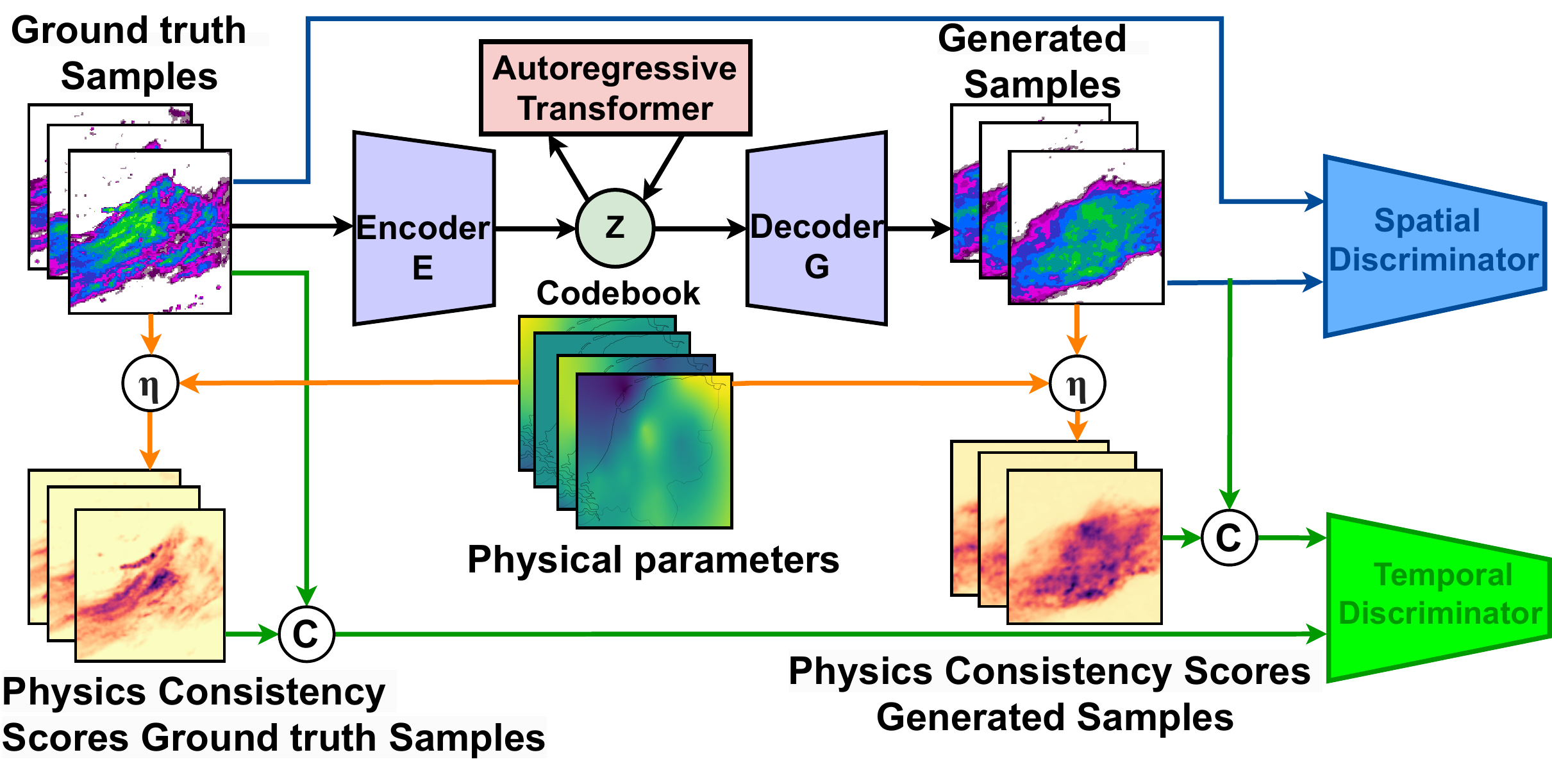}
\caption{The proposed PID-GAN model structure. C means concatenation and $ \eta $ represent $ \eta_k = e^{-\lambda \mathcal{R}^{(k)}(x,\hat{x})}$, refering to the equation of the physics consistency score.}
\label{fig:Model}
\end{figure*}
\subsubsection{Generator}
The generator in our model incorporates a deep generative model with two components: Vector Quantization Generative Adversarial Network (VQ-GAN) \cite{esser2021taming}, that learns the mapping between precipitation maps and discrete tokens, and an Autoregressive Transformer (AT) \cite{vaswani2017attention} that models the dynamics between tokens of consecutive timesteps. \newline
\textbf{VQ-GAN}:
The architecture of the VQ-GAN model consists of a CNN encoder ($E$) and decoder ($G$), a codebook ($\mathcal{Z}$), which define a VQ-VAE \cite{van2017neural} and a patch-based discriminator \cite{isola2017image}, indicated as spatial discriminator ($D$). 
The VQ-VAE is trained by optimizing:
\begin{equation}
\begin{aligned}
   \mathcal{L}_{\text{vq-vae}} = & \left\| x - \hat{x} \right\|_1 + \left\| sg[E(x)] - z_q \right\|_2^2 \\
   & + \left\| sg[z_q] - E(x) \right\|_2^2 + \mathcal{L}_{\text{perceptual}}(x, \hat{x}).
\end{aligned}
\label{eq: VQ-GAN loss}
\end{equation}
Here, $sg[.]$ represents the stop-gradient operation, which prevents back-propagating gradients. The loss function comprises four terms: the reconstruction loss $\mathcal{L}_{\text{rec}}= \left\| x - \hat{x} \right\|_1$, comparing the original input $x$ (radar maps) with its reconstruction $\hat{x}$ (reconstructed maps). The commitment loss, covered by the second and third terms, penalizes discrepancies between the encoded representations and codebook entries and optimizes the codes within the codebook.  The fourth term, perceptual loss, assesses high-level semantic differences between $x$ and $\hat{x}$ \cite{esser2021taming}. The VQ-GAN is optimized training adversarially VQ-VAE, which acts as a generator, and a  patch-based discriminator (i.e., spatial discriminator), using the following loss function:
\small
\begin{equation}
    \underset{E,G,\mathcal{Z}}{\mathrm{argmin}} \max_D \mathbb{E}_{x \sim p(x)} \left[ \mathcal{L}_{\text{vq-vae}}(E,G,\mathcal{Z}) + \lambda \mathcal{L}_{\text{GAN}}(\{E,G,\mathcal{Z}\}, D)\right],
\label{eq: VQ-GAN1 loss}
\end{equation}
\begin{equation}
   \label{eq: VQ-GAN AD loss}
\mathcal{L}_{\text{GAN}}(\{E, G, \mathcal{Z}\}, D) = \left[ \log D(x) + \log(1 - D(\hat{x})) \right]
\end{equation}
\begin{equation}
   \label{eq: adaptive weight}
\lambda_{\text{GAN}} = \frac{\nabla_{G} \left[\mathcal{L}_{rec}\right]}{\nabla_{G} \left[\mathcal{L}_{GAN}\right] + \delta}.
\end{equation}
\normalsize
Here, $\mathcal{L}_{\text{GAN}}( \{ E, G, \mathcal{Z} \}, D)$ represents the discriminator loss and $\lambda_{\text{GAN}}$ is the adaptive weight determined by $\nabla_{G}[.]$, which represents the gradient of the input concerning the final layer of decoder. $\delta=10^{-6}$ is a scalar for numerical stability.\newline
\textbf{Autoregressive Transformer}: The AT architecture aims to model the dynamics between consecutive precipitation maps \cite{vaswani2017attention}. The ground truth precipitation maps are quantized into $\mathbf{z}_q = q(\mathbf{E}(x))$, producing a sequence $\mathbf{s} \in \{0, \ldots, |\mathbf{Z}|-1\}^{h \times w}$, representing VQ-GAN codebook indices. These indices are transformed into continuous vectors by an embedder and augmented with positional embeddings to provide order information. The transformer then processes these vectors, with the head module refining the output into logits, which represent the probability of using a specific token. These logits are used to compute a cross-entropy loss that compares predicted token probabilities with the actual tokens:
\begin{equation}
\label{eq:CEPLoss}
\mathcal{L}_{\text{Transformer}} = \mathbb{E}_{x\sim p(x)}[-\log \prod_{i=1}^{N} p(\mathbf{s}_i | \mathbf{s}_{<i})].
\end{equation}
Given a sequence of indices $ \mathbf{s}_{<i}$, the transformer is trained to predict the distribution of the consecutive indices $\mathbf{s}_i$.
The AT employs a causal attention mechanism which accesses only previously seen and current tokens when predicting the next one in a sequence, enabling efficient and context-sensitive output production. 
\subsection{Physics-Informed Discriminator: PID-GAN}
Following the PID-GAN \cite{daw2021pid} framework, we use physics residuals to compute a physics consistency score ($\eta$) for each prediction, indicating the likelihood of the prediction being physically consistent. These physics consistency scores are fed into a temporal discriminator \cite{jing2019aenn} as additional inputs, such that it distinguishes between real and fake sequences of precipitation maps by learning from the underlying distribution of labelled points and using the additional physics supervision. \newline
\textbf{Estimating Physics Consistency Scores:} Formally, we compute the physics consistency score of a prediction $\hat{x}$ for the $k$-th physical constraints as $ \eta_k = e^{-\lambda \mathcal{R}^{(k)}(x,\hat{x})}$,
where $\mathcal{R}^{(k)}$ represents the physical equation used to describe the phenomena. The larger $\eta_k$, the more prediction $\hat{x}$ obeys the $k$-th physical constraint. Following \cite{daw2021pid}, the temporal physics-informed discriminator is trained by optimizing: 
\begin{equation}
\label{eq: PID D loss}
\mathcal{L}_D(\phi) = - \frac{1}{N} \sum_{i=1}^{N} \log (D(x_{i}, \eta_{i})) - \frac{1}{N} \sum_{i=1}^{N} \log (1 - D(\hat{x}_{i}, \hat{\eta}_{i})).
\end{equation}
Here, $\eta_{i}$ and $\hat{\eta}_{i}$ represent the physics consistency score for ground truth and generated data, ensuring alignment with physical laws.
\newline
\textbf{Moisture Conservation Equation:}
The Moisture Conservation Equation \cite{ rohli2021meteorology, georgakakos1984hydrologically} in NWP, which describes the relationship between atmospheric moisture content, evaporation, and precipitation (P), is given by:
\begin{equation}
\label{eq:Moisture_Conservation with Makkink equation}
    \frac{\partial q}{\partial t} = -u \frac{\partial q}{\partial x} - v \frac{\partial q}{\partial y} - \omega \frac{\partial q}{\partial z} +ET - P
\end{equation}
\begin{equation}
\label{eq:Makkink equation}
    ET = 0.65 \frac{\Delta}{\Delta+\gamma}\frac{R_s}{\lambda}.
\end{equation}
Here, $u$ is the west-to-east wind component ($m/s$), $v$ is the south-to-north wind component ($m/s$), $\omega$ is the vertical wind component ($m/s$), $q$ is the specific humidity ($ g\ g^{-1}$), $ET$ is evapotranspiration rate ($mm/h$), $\Delta$ is the derivative with respect to temperature of the saturation vapour pressure ($Pa/^\circ\text{C}$), $\gamma$ is the psychrometric constant ($J/m^2 $), $\lambda$ is the latent heat of vaporization ($J/g$), and $R_s$ is global radiation ($Pa/^\circ\text{C}$)
. The evapotranspiration can be estimated using the Makkink Equation \eqref{eq:Makkink equation} \cite{de1987penman}, which relies on temperature and solar radiation data, yielding relatively accurate results in cold and temperate humid climates. Measuring vertical wind speed $\omega$ poses challenges due to its low intensity and high spatial variability, the requirement for sensitive equipment, atmospheric stability, and the high costs and complexity of accurate measurements, resulting in a lack of datasets with these measurements. Omitting the term $\omega \frac{\partial q}{\partial z}$ simplifies the process but risks overlooking crucial moisture transport between atmospheric layers, potentially impacting the balance of moisture within the atmosphere's three-dimensional dynamics. To compensate for the lack of vertical wind speed data, the model shifts focus to the horizontal wind components (U and V) at different elevation levels, using the ERA5 dataset for wind measurements at $100$ meters ($u_{100}$ and $v_{100}$) and AWS data at $10$ meters ($u_{10}$ and $v_{10}$). It's important to note that atmospheric interactions, which significantly influence weather patterns, extend up to the end of the troposphere, approximately 10 km in altitude. Therefore, by focusing on wind speed within the lower 100 meters, the approach inevitably entails a degree of uncertainty, given the comprehensive atmospheric interactions occurring beyond this range. By integrating these horizontal components from both 10-meter and 100-meter altitudes, the model adapts to variations in altitude for atmospheric moisture analysis. This approach simplifies equation \eqref{eq:Moisture_Conservation with Makkink equation} to approximate moisture transport across altitudes without needing direct vertical wind speed measurements, offering a more detailed view of the atmosphere's moisture dynamics. As a result, we can  define the physical constraint for the proposed PID-GAN as:
\small
\begin{equation}
\label{eq: constrain}
\mathcal{R}_q = -\frac{\partial q}{\partial t} -u_{10} \frac{\partial q}{\partial x} - v_{10} \frac{\partial q}{\partial y} -u_{100} \frac{\partial q}{\partial x} - v_{100} \frac{\partial q}{\partial y} + ET - P.
\end{equation}
\normalsize
where the equation is computed at the pixel level. 
\section{Experiments}
\begin{table*}[htbp]
\caption{3-hour averaged precipitation nowcasting skill of different models (Pixel-level evaluation). Top and second-best performances are highlighted in bold and underlined, respectively. PID-GAN(-P) represents the proposed model without physical constraints, and PID-GAN(-PT) represents the proposed model without physical constraints and temporal discriminator.}
\centering
\resizebox{\textwidth}{!}{
\begin{tabular}{|c|c|c|c|c|c||c|c|c|}
\hline
 & \textbf{PySTEPs} & \textbf{TECO} & \textbf{Nuwä-EVL} & \textbf{NowcastingGPT+EVL} & \textbf{NowcastingGPT} & \textbf{PID-GAN} & \textbf{PID-GAN(-P)} & \textbf{PID-GAN(-PT)} \\ \hline
\textbf{PCC $(\uparrow)$} & 0.219 & 0.149 & 0.202 & 0.253 & 0.241 & \textbf{0.313} & \underline{0.288} & 0.250 \\ \hline
\textbf{MAE $(\downarrow)$} & 0.798 & \textbf{0.664} & 0.938 & 0.714 & 0.725 & \underline{0.686} & 0.706 & 0.692 \\ \hline
\textbf{MSE $(\downarrow)$} & 4.210 & 3.335 & 3.592 & 3.298 & 3.293 & \textbf{3.117} & \underline{3.162} & 3.271 \\ \hline
\textbf{CSI(1mm) $(\uparrow)$} & 0.250 & 0.097 & 0.262 & 0.267 & 0.21 & \textbf{0.313} & \underline{0.296} & 0.234 \\ \hline
\textbf{CSI(8mm) $(\uparrow)$} & 0.008 & 0.001 & 0.006 & \underline{0.009} & 0.005 & \textbf{0.011} & 0.008 & 0.004 \\ \hline
\textbf{FAR(1mm) $(\downarrow)$} & 0.617 & 0.662 & 0.623 & 0.587 & \underline{0.579} & 0.583 & 0.601 & \textbf{0.549} \\ \hline
\textbf{FAR(8mm) $(\downarrow)$} & 0.592 & \textbf{0.361} & \underline{0.399} & 0.502 & 0.513 & 0.529 & 0.499 & 0.435 \\ \hline
\textbf{FSS(1km) $(\uparrow)$} & 0.375 & 0.163 & 0.394 & \underline{0.432} & 0.414 & \textbf{0.451} & 0.430 & 0.428 \\ \hline
\textbf{FSS(10km) $(\uparrow)$} & 0.467 & 0.211 & 0.456 & 0.493 & 0.463 & \textbf{0.534} & \underline{0.510} & 0.481 \\ \hline
\textbf{FSS(20km) $(\uparrow)$} & 0.522 & 0.248 & 0.498 & 0.534 & 0.508 & \textbf{0.591} & \underline{0.565} & 0.521 \\ \hline
\textbf{AUC $(\uparrow)$} & 0.454 & 0.378 & 0.516 & \underline{0.538} & 0.510 & \textbf{0.567} & 0.532 & 0.520 \\ \hline
\end{tabular}
}
\label{tab:forecasting-models-metrics}
\end{table*}
Our investigation addresses two main research questions: (a) Can physical supervision improve the accuracy of precipitation nowcasting? (b) Does incorporating physical data into the model's design enhance its ability to detect extreme precipitation events? 
We compare the proposed model to a classic benchmark, namely Pysteps \cite{pulkkinen2019pysteps}, a temporally consistent video prediction benchmark, TECO \cite{yan2022temporally}, and two benchmarks that use Extreme Value loss regularization, namely  Nuwä-EVL \cite{bi2023nowcasting} and NowcastingGPT+EVL \cite{Meo2024extreme}. The nowcasting task is configured with parameters set to $N=3$ conditioning timesteps and $M=6$ predicted maps, with each timestep representing a realistic scenario of 30 minutes. Outputs from all models are the average ensembles of five sample predictions. The PySteps model employed in this study adheres to the probabilistic configuration parameters outlined in \cite{imhoff2020spatial}.

To quantitatively assess the predictions we calculate visual fidelity metrics including Mean Squared and Absolute Errors (MSE, MAE) and Pearson Correlation Score (PCC), and nowcasting metrics, such as Critical Success Index (CSI), False Alarm Ratio (FAR) and Fractional Skill Score (FSS). Precipitation thresholds of 1 and 8 mm are set for CSI and FAR, and FSS is evaluated at spatial scales of 1, 10, and 20 km with a 1 mm threshold. Since fidelity metrics cannot capture extreme event classification, we plot a precision-recall curve of the extremes to assess the considered baselines in terms of extreme classification capabilities. 
\begin{figure}[htbp]
\centering
\includegraphics[width = 0.81\linewidth]{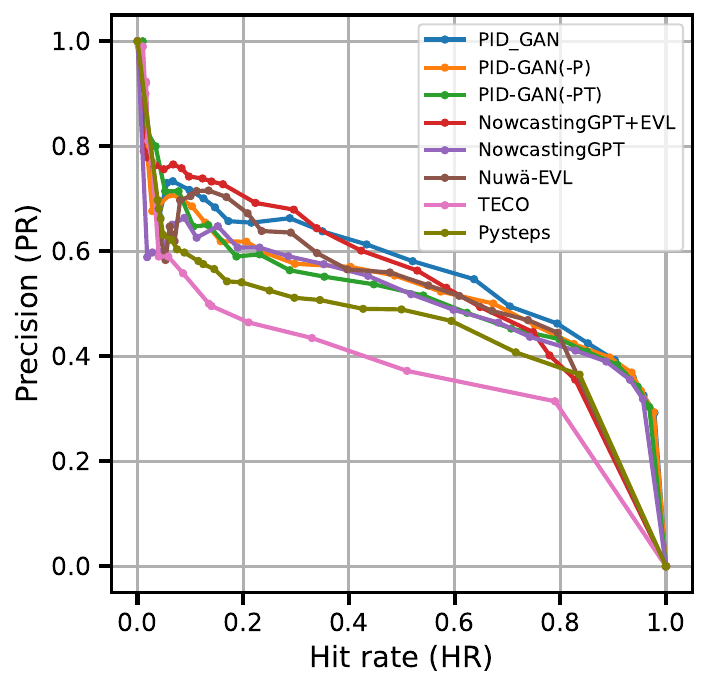}
\caption{The precision-recall curves for detecting extreme
 events over 3 hours at 12 Dutch catchments. Every point from right to left represents a different precipitation threshold $(0.5$ to $10 \text{mm}/3\text{h})$ for prediction and a fixed threshold for ground truth by definition of extreme events\cite{bi2023nowcasting}. } 
\label{fig:PR_curves}
\end{figure}
Table \ref{tab:forecasting-models-metrics} presents a comparison of nowcasting downstream performances for the validated models across the entire study area. The PID-GAN model's advancements are particularly notable in Mean Squared Error (MSE) and Pearson Correlation Score (PCC), indicating a strong correlation with actual precipitation events and a high degree of prediction accuracy. Additionally, the model excelled in the Fractional Skill Score (FSS) across different spatial scales and significantly outperformed all benchmarks in the Critical Success Index (CSI) at both light (1mm) and heavy (8mm) precipitation thresholds. This performance suggests an enhanced ability to detect and accurately predict a wide range of precipitation events, from light showers to severe storms. 

Figure \ref{fig:PR_curves} highlights the PID-GAN model's outstanding performance in detecting extreme precipitation events within 12 Dutch catchments \cite{imhoff2020spatial}, as illustrated by its precision-recall curves. These curves not only underscore the model's superior capability in forecasting extreme events when compared to all benchmarks but also reveal PID-GAN's exceptional balance between precision and recall, evidenced by achieving the highest area under the Precision-Recall curve (AUC). This is particularly noteworthy in the context of effective nowcasting, where identifying extreme weather patterns is crucial. Additionally, the analysis reveals a significant insight: removing physical constraints from the PID-GAN model (PID-GAN(-P)) leads to a decrease of $6.17\%$ in AUC. This drop highlights the critical role that physical constraints play in enhancing the accuracy of precipitation map predictions, further validating the importance of integrating physical data into the model's design for improved performance.

\section{Conclusion}
In conclusion, the proposed PID-GAN model has demonstrated significant effectiveness in nowcasting precipitation and accurately predicting extreme precipitation events. Addressing a notable challenge in nowcasting, the model not only outperforms existing benchmarks in terms of accuracy but also significantly enhances forecast precision. This study innovates by integrating the moisture conservation equation as a physical constraint, offering a novel approach to improving forecast accuracy. Future research will aim to further enhance the model's capabilities by incorporating additional physical constraints, such as the impact of air temperature on extreme precipitation events, and by evaluating the model's performance across diverse geographical settings.

\bibliographystyle{IEEEtran}
\bibliography{refs} 

\end{document}